\documentclass[journal]{IEEEtran}
\usepackage{amsmath,amsfonts}
\usepackage{algorithmic}
\usepackage{algorithm}
\usepackage{array}
\usepackage[caption=false,font=normalsize,labelfont=sf,textfont=sf]{subfig}
\usepackage{textcomp}
\usepackage{stfloats}
\usepackage{url}
\usepackage{verbatim}
\usepackage{graphicx}
\usepackage{cite}

\begin{document}
\title{Dockformer: A transformer-based molecular docking paradigm for large-scale virtual screening}
\author{Zhangfan Yang, Junkai Ji, Shan He, Jianqiang Li, Tiantian He, \IEEEmembership{Member,~IEEE}, Ruibin Bai, \IEEEmembership{Senior Member,~IEEE}, Zexuan Zhu, \IEEEmembership{Senior Member,~IEEE}, Yew Soon Ong, \IEEEmembership{Fellow, ~IEEE}
\thanks{Zhangfan Yang and Ruibin Bai are with the School of Computer Science, University of Nottingham Ningbo, Ningbo, 315199, China. (email: 2070276085@email.szu.edu.cn; ruibin.bai@nottingham.edu.cn).}
\thanks{Junkai Ji, Jianqiang Li and Zexuan Zhu are with the National Engineering Laboratory for Big Data System Computing Technology, Shenzhen University, Shenzhen, 518060, China (email: jijunkai@szu.edu.cn; lijq@szu.edu.cn; zhuzx@szu.edu.cn).}
\thanks{Tiantian He is with the Center for Frontier AI Research, Institute of High Performance Computing, Singapore Institute of Manufacturing Technology, Agency for Science, Technology and Research (A*STAR), Singapore 138632 (email: he\_tiantian@ihpc.a-star.edu.sg).}
\thanks{Shan He is with the School of Computer Science, University of Birmingham, Birmingham, B15 2TT, UK (email: s.he@cs.bham.ac.uk).}
\thanks{Yew Soon Ong is with the School of Computer Science and Engineering, Nanyang Technological University, Nanyang Avenue, 639798, Singapore (email: asysong@ntu.edu.sg).}

}


\maketitle

\begin{abstract}
Molecular docking is a crucial step in drug development, which enables the virtual screening of compound libraries to identify potential ligands that target proteins of interest. However, the computational complexity of traditional docking models increases as the size of the compound library increases. Recently, deep learning algorithms can provide data-driven research and development models to increase the speed of the docking process. Unfortunately, few models can achieve superior screening performance compared to that of traditional models. Therefore, a novel deep learning-based docking approach named Dockformer is introduced in this study. Dockformer leverages multimodal information to capture the geometric topology and structural knowledge of molecules and can directly generate binding conformations with the corresponding confidence measures in an end-to-end manner. The experimental results show that Dockformer achieves success rates of 90.53\% and 82.71\% on the PDBbind core set and PoseBusters benchmarks, respectively, and more than a 100-fold increase in the inference process speed, outperforming almost all state-of-the-art docking methods. In addition, the ability of Dockformer to identify the main protease inhibitors of coronaviruses is demonstrated in a real-world virtual screening scenario. Considering its high docking accuracy and screening efficiency, Dockformer can be regarded as a powerful and robust tool in the field of drug design.
\end{abstract}


\section{Introduction}
\IEEEPARstart{I}{n} drug discovery, identifying the candidate compounds that target biological macromolecules remains challenging because of the long development time and expensive wet-laboratory experiments. Virtual screening using molecular docking approaches can significantly improve the initial hit rate of drug candidates with great diversity and high binding affinity \cite{lyu2019ultra, gorgulla2020open}. Recently, the number of synthesizable molecules in make-on-demand libraries has expanded from 3.5 million to 29 billion. The docking performance can steadily improve as the library size increases \cite{lyu2023modeling}. However, in large-scale virtual screening (LSVS) tasks, the computational cost and time of docking methods become major challenges that most researchers in academia and industry cannot overcome \cite{sadybekov2023computational}.

Traditional docking approaches use scoring functions to measure the binding affinity of a given protein--ligand complex and then find the best binding conformation by applying optimization algorithms to minimize these functions \cite{li2019overview}. For example, GOLD uses a genetic algorithm to search complex conformations \cite{verdonk2003improved}, and AutoDock combines a genetic algorithm with a simulated annealing algorithm \cite{morris2009autodock4}. Although these optimization-based docking methods are commonly used in modern drug designs because of their good usability and interpretability, they still face the following challenges: the scoring functions are generally not precise enough, and optimization algorithms cannot guarantee that the global optimum is found every time. Although several advanced methods can offer reliable binding affinity predictions \cite{Co-VAE, GNN_pre}, docking approaches require multiple independent optimization processes to sample possible binding conformations for each protein--ligand pair, leading to very high computational costs in LSVS tasks \cite{gorgulla2020open, yu2024survey}.

Inspired by the groundbreaking advancement of AlphaFold2 in protein structure prediction \cite{jumper2021highly}, a series of deep learning (DL)-based methods have emerged to solve molecular docking tasks \cite{buttenschoen2024posebusters}. These approaches can be divided into three categories according to their neural network architectures: graph neural networks (GNNs)-based \cite{stark2022equibind}, transformer-based \cite{zhou2023unimol} and diffusion model-based docking methods \cite{corso2023diffdock}. The primary motivation of these studies is twofold: first, improving the ligand docking accuracy with the aid of the powerful learning capabilities of DL technologies, and second, speeding up the screening process by directly predicting ligand binding conformations to skip the time-consuming optimization of traditional docking approaches \cite{fadahunsi2024revolutionizing}. Although tremendous efforts have been made to develop DL-based docking tools, few can perform well in docking accuracy and screening speed simultaneously due to inadequate generalizability and non-end-to-end architectures \cite{isert2023structure, yu2023deep}. In addition, existing methods solely focus on 1D sequential, 2D graph topological or 3D structural in isolation, and fail to leverage the integration and complementarity of each modality for capturing the inherent interactions between proteins and ligands. Therefore, how to use DL models to generate protein-ligand binding conformations precisely and efficiently is still an open question in LSVS tasks.

In this study, a novel transformer-based architecture named Dockformer is proposed to overcome the above-mentioned issues of current DL-based docking methods. Specifically, Dockformer uses two separate encoders to leverage multimodal information to generate latent embeddings of proteins and ligands and can thus effectively capture molecular geometric details, including 2D graph topology and 3D structural knowledge. A binding module is then employed to detect intermolecular relationships effectively on the basis of learned latent embeddings. Finally, in the structure module, the established relationships are utilized to generate the complex conformations directly, and the coordinates of the ligand atoms are calculated in an end-to-end manner. In addition, the corresponding confidence measures of each generated conformation are utilized to distinguish binding strengths instead of traditional scoring functions. In summary, distinct from conventional DL-based and optimization-based docking methods, the multimodal information fusion equips Dockformer with superior docking accuracy, and the end-to-end architecture enables it to simultaneously speed up the conformation generation process by orders of magnitude. Thus, this method can meet the rapid throughput requirements of LSVS tasks. Dockformer, as a robust and reliable protein-ligand docking approach, may significantly reduce the development cycle and cost of drug design.

The remainder of this paper is organized as follows: Section
\ref{related_works} introduces related works in molecular docking. In Section \ref{methods}, the architecture details of Dockformer are presented. Section \ref{experiment} analyzes docking performance and utilizes confidence metrics for large-scale virtual screening, while discussing optimization algorithms for physical plausibility. Finally, Section \ref{conclusion} reviews the use of AI technologies for screening large-scale compound libraries and discusses the potential of de novo drug design using generative models and deep reinforcement learning to streamline the screening process.

\section{Related work}
\label{related_works}
DL-based molecular docking methods can be divided into three main categories: GNNs-based, transformer-based and diffusion-based methods.
The models in the first class encode proteins and ligands as graphs and use equivariant GNNs to predict intermolecular binding interactions \cite{2023zhang}. For instance, DeepDock utilizes GNNs to construct a mixture density network, which is based on the distance likelihood of ligand-target node pairs and can act as a scoring function \cite{mendez2021geometric}. Then, DeepDock can accurately search complex conformations by optimizing the scoring function. EquiBind employs a SE(3)-equivariant GNN to detect the interactions between protein residues and ligand atoms, and uses gradient descent algorithms to determine the translation, rotation and torsion of binding conformations \cite{stark2022equibind}. Similarly, TankBind uses a trigonometry-aware GNN to predict protein-ligand intermolecular distances. Then, it adopts a multi-dimensional scaling method to reconstruct the ligand atom coordinates based on the pair distances \cite{lu2022tankbind}. KarmaDock combines the methodologies of DeepDock and EquiBind, which utilizes a graph transformer neural network to learn pair distance distributions and employs E(n)-equivariant GNNs to generate binding conformations directly \cite{zhang2023efficient}. For molecular docking tasks, graph models can directly handle the structural geometry and effectively process the symmetry properties of molecular representations, enabling the movement direction and amplitude of ligand atoms to be updated in each message passing iteration. However, the over-smoothing issue results in the inadequate generalizability of these GNN-based methods. The performance of the methods has not yet reached that of conventional docking methods \cite{isert2023structure}.

The models in the second class are based on transformer architectures, which can efficiently capture long-range dependencies among intra- and intermolecular tokens. For example, Uni-Mol has pioneered the use of atom and pair representations to encode ligands and protein pockets, and employed the self-attention layers with pair biases to share information between each representation. It generates the coordinates of ligand atoms via two distance matrices predicted by pairwise representations \cite{zhou2023unimol}. Inspired by Alphafold2, GAABind incorporates the additional triangular self-attention layers into the main architecture of Uni-Mol, which can capture the geometric and topological properties of binding pockets and ligands \cite{tan2024gaabind}. In addition, considering transformer models typically exist urgent requirements of enormous training data, CarsiDock and HelixDock customized large-scale complex structure datasets for pretraining and used the crystallized structure dataset for fine-tuning to improve their generalization abilities \cite{cai2024carsidock, liu2023pre}. Although these transformer-based models can achieve satisfactory docking accuracy, most still require independent optimization procedures to generate binding poses from the predicted interaction distance maps or distributions. Therefore, their inference processes remain time-consuming and are expensive for LSVS tasks \cite{yu2023deep}. Furthermore, these transformer-based methods do not adequately account for the positional embeddings of individual tokens (atoms), which ultimately compromises the model's generalizability and overall performance. In addition, these models output physically implausible conformations with steric clashes and incorrect bond lengths and angles since they ignore the essential topological information of molecules during the conformation generation process \cite{buttenschoen2024posebusters}.

Unlike the aforementioned models, which treat molecular docking as regression problems, models in the third class frame docking problems as generative modeling tasks. Specifically, DiffDock uses a denoising diffusion probabilistic model over the non-Euclidean manifold of ligand conformations, and then maps the manifold to predict the translation, rotation and torsion of ligands \cite{corso2023diffdock}. DynamicBind adopts an equivariant geometric diffusion network to construct a smooth energy landscape, which can be used to recover ligand conformations based on the unbound structures of proteins \cite{lu2024dynamicbind}. Furthermore, NeuralPLexer employed a diffusion-based generative model to predict complex structures by solely inputting protein sequences and ligand molecular graphs \cite{qiao2024state}. AlphaFold3, which is recently proposed, applies diffusion transformer models as decoders to simultaneously calculate each atom coordinate of proteins and ligands and yields very remarkable prediction accuracy in molecular docking tasks \cite{abramson2024accurate}. However, such generative models require the sampling of many noisy conformations for denoising step by step, leading to very high computational complexity and slow docking speed. Despite the superior prediction performance of AlphaFold3, its inference time is longer than that of most docking approaches, making it incapable of virtually screening the billions of compounds in large-scale libraries.

To overcome the drawbacks of conventional DL-based docking algorithms, Dockformer first uses two separate encoders to integrate multimodal information for generating latent embeddings of proteins and ligands. Each encoder effectively captures molecular geometric details from 2D graph topology and 3D structural knowledge, enabling a more comprehensive understanding of molecular interactions. It is because 2D graph information allows us to grasp the bonding relationships and connectivity patterns that are crucial for accurate docking predictions, and the 3D structural information provides spatial context, ensuring an account for the actual conformational geometry of molecules. Second, Dockformer uses an end-to-end decoder to generate the complex conformations and the corresponding confidence measures directly. It can skip time-consuming optimization and denoising processes, thereby significantly accelerating the docking procedure and enhancing computational efficiency. The confidence measures can be used to distinguish binding strengths, replacing traditional scoring functions that may not accurately reflect true binding affinities. These properties empower Dockformer to achieve superior docking accuracy and screening efficiency in the LSVS tasks, compared with state-of-the-art DL-based algorithms.

\begin{figure*}[t]
\centering
\includegraphics[width=0.9\linewidth]{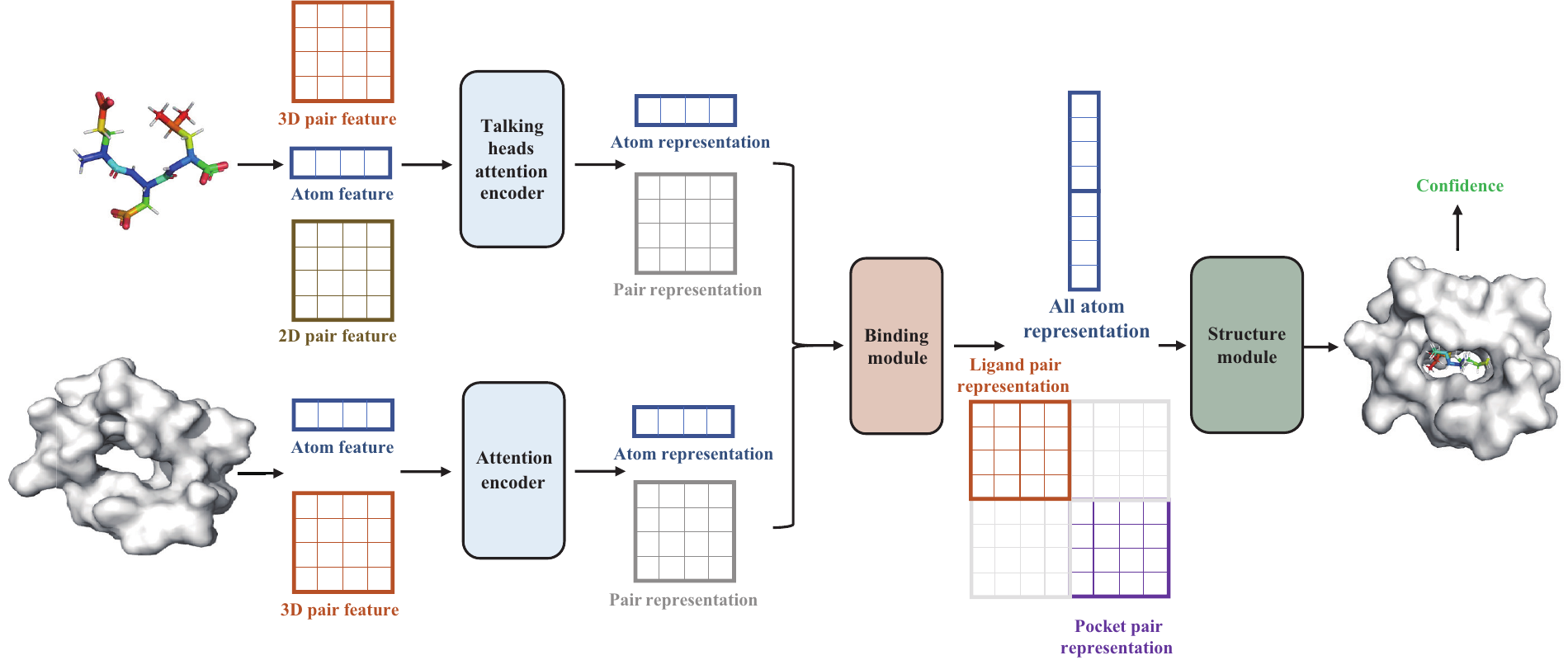}
\caption{Network architecture of Dockformer, constituted by two independent encoders, a binding module and an end-to-end structure module.}
\label{fig:architecture}
\end{figure*}

\section{Dockformer}
\label{methods}
\subsection{Architecture Overview}
The essence of molecular docking lies in detecting nonbonded interactions between the atoms of a ligand and a protein pocket. The protein is conventionally regarded as a rigid body to simplify the calculations, and docking algorithms aim to generate the binding conformation of ligands within the protein--ligand complex on the basis of the predicted atom interactions. Consequently, the proposed Dockformer algorithm is designed to directly predict the 3D coordinates of all heavy atoms of the ligands for a given protein pocket. The architecture of the proposed Dockformer is depicted in Fig. \ref{fig:architecture}. Dockformer consists of three stacked primary blocks. First, two independent encoders are used to encode the multimodal information of the ligand and binding pocket, and intramolecular interactions are exploited to produce their intrinsic representations. Second, a binding module captures intermolecular interactions between the binding pocket and ligand to generate the corresponding latent embeddings. Finally, the latent embeddings are fed into the structure module to predict the binding conformation of the ligand by considering the precise 3D coordinates of each atom.

\subsection{Featurization Methodology}
The network architecture simultaneously incorporates the 1D sequence, 2D graph and 3D geometry information of the ligand and protein pocket as inputs, enabling valuable insights from distinct modalities. Let $A^0 = [\mathrm{A}_1^0, \mathrm{A}_2^0, ..., \mathrm{A}_N^0]$ denote the initial atom features used to encode the sequence information, where $N$ is the number of heavy atoms and $\mathrm{A}_n^0$ represents the atom type of the $n$-th atom by using a one-hot encoding scheme. Additionally, 2D graph information is encoded as the chemical bonds and structural interconnections between atom pairs in the ligand. Two-dimensional graph pair features $\mathit{\Phi}_{ij}^\mathit{2D}$ contain two representations. First, $\mathit{\Phi}^\mathit{SPD}_{ij}$ represents the connection feature, which uses the shortest path distance between atoms $i$ and $j$ to reflect their connection relation in the graph. Second, $\mathit{\Phi^{edge}}$ records the edge feature to reflect the bond information. Denoting the edges along the shortest path of atoms $i$ and $j$ as $\mathrm{E}_{ij}=(\mathrm{e}_1,\mathrm{e}_2,...,\mathrm{e}_N)$, the edge feature can be calculated by $\mathit{\Phi}^{\mathit{edge}}_{ij} = \frac{1}{N}\sum_{n=1}^N\mathrm{e}_n(w_{edge})^T$, where $w_{edge}$ are the learnable parameters. Notably, both $\mathit{\Phi}^\mathit{SPD}$ and $\mathit{\Phi}^\mathit{edge} \in \mathbb{R}^{N \times N}$ need to be calculated for the protein pocket, which is considered rigid during the docking process.

\begin{figure*}[t]
\centering
\includegraphics[width=0.9\linewidth]{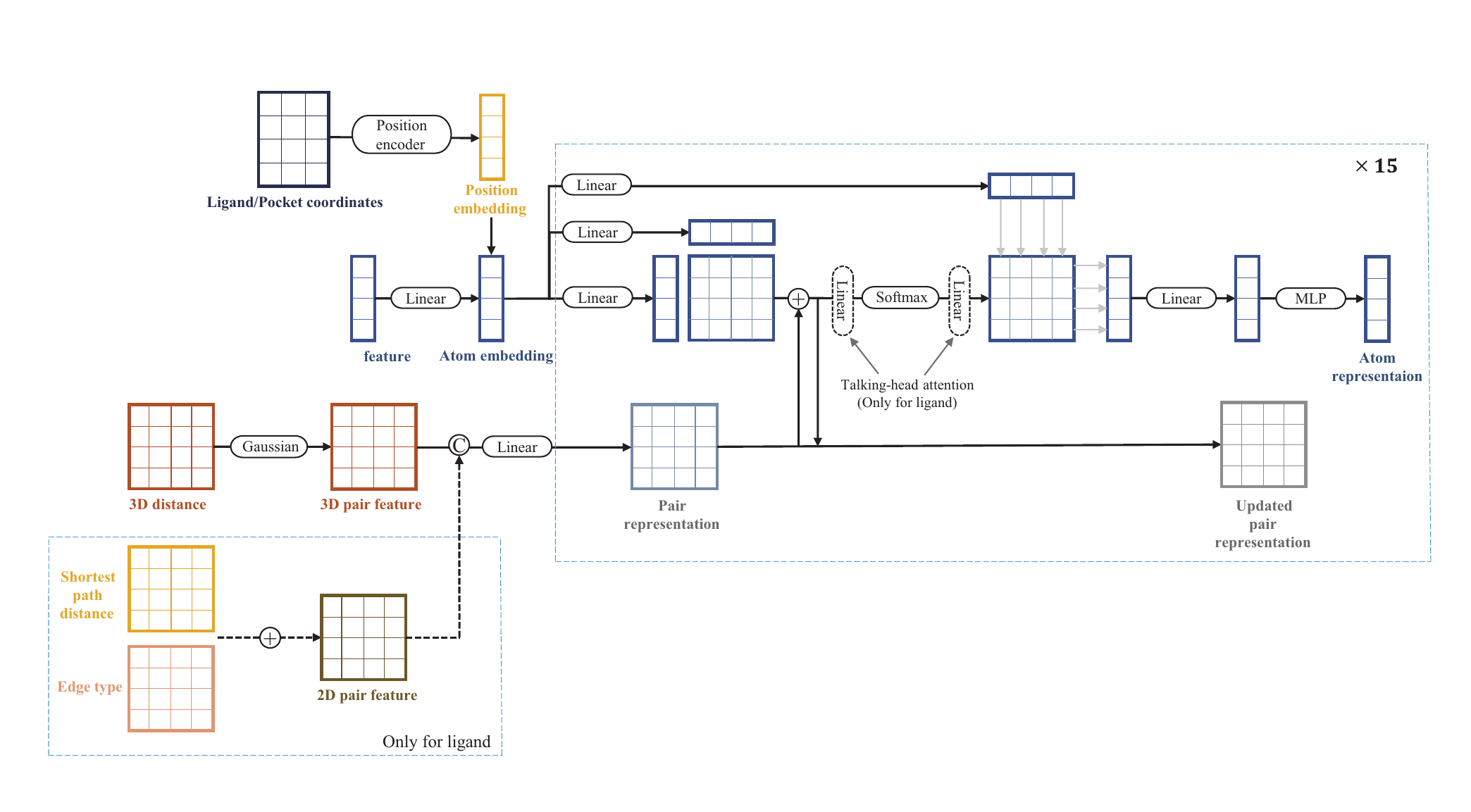}
\caption{Encoder modules are used to encode the input representations of protein pockets and ligands, where features of 2D graph topology and 3D structural knowledge are integrated by modified self-attention layers to generate latent embeddings of molecules.}
\label{fig:encoder}
\end{figure*}

Next, the 3D geometric information of the ligand and protein pocket is encoded as two representations. Regarding the atoms as point clouds, the first representation is the learnable global position embedding $GPE_i$ to reflect spatial information according to the 3D coordinate $\{\mathrm{P}_i^{x}, \mathrm{P}_i^{y}, \mathrm{P}_i^{z}\}$ of each atom, where $i \in \{1, 2, ..., I\}$ and $I$ denotes the dimensional number. $GPE_n^i$ of the $n$-th atom can be calculated by
\begin{equation}
GPE_n^i = \mathrm{MLP}\Big(\mathrm{Concat}\big(sine(\mathrm{P}_i^{x}), sine(\mathrm{P}_i^{y}), sine(\mathrm{P}_i^{z})\big)\Big),
\label{eq7}
\end{equation}
where $sine$ uses sine and cosine functions to map each coordinate to a required position vector. These vectors are subsequently concatenated to generate a global position vector whose dimension is subsequently reduced to match the embedding dimension via a shadow multilayer perceptron (MLP). The calculation of $sine$ can be described by
\begin{equation}
\begin{aligned}
sine{(\mathrm{P}_i, i)} =
\begin{cases}
\sin(\mathrm{P}_i/10000^{2i/I}),\ \mathrm{if} \ i \ \mathrm{is \ even};\\
\cos(\mathrm{P}_i/10000^{2i/I}), \ \mathrm{if} \ i \ \mathrm{is \ odd}.
\end{cases}
\end{aligned}
\label{eq8}
\end{equation}

The second representation is the 3D pair feature $\mathit{\Phi}_{ij}^\mathit{3D}$ to encode the geometric relation. The interatomic distance $d_{ij}$ between each atom pair $i$ and $j$ is calculated. Each element is encoded by $K$ Gaussian basic kernel functions $\mathcal{N}(\hat{d}_{ij};\mu_k,\sigma_k) =\frac{1}{\sqrt{2\pi\sigma_k}}\exp{(-\frac{1}{2\pi\sigma_k^2}(\hat{d}_{ij}-\mu_k)^2)}$, where $ k \in\{1,...,K\}$, $\mu_k$ and $\sigma_k$ represent the predefined mean and the standard deviation, respectively. The transformed distance is $\hat{d}_{ij} =  u_{ij} \cdot d_{ij} + v_{ij}$, where $u_{ij}$ and $v_{ij}$ are learnable parameters that share values for pairs of the same atom types. Finally, $\mathit{\Phi}_{ij}^\mathit{3D}$ can be obtained via the nonlinear transformation of $\mathcal{N}(\hat{d}_{ij})$, described by
\begin{equation}
\mathit{\Phi}_{ij}^\mathit{3D} = \mathrm{LeakyReLU}\big(\mathcal{N}(\hat{d}_{ij})W_1^{3D}\big)W_2^{3D},
\end{equation}
where $W_1^{3D} \in \mathbb{R}^{K \times K}$ and $W_2^{3D} \in \mathbb{R}^{K \times 1}$ are the weights of the linear transformations and $\mathrm{LeakyReLU}$ is the activation function. The featurization methodology can effectively capture the intricacies and diversities of molecular structures, thereby enhancing the performance and generalization capabilities of the proposed model.

\subsection{Encoder Modules}
Two encoders are used to update the representations of both the ligand and protein pockets, which share the same architecture but have different weights. The architecture of the encoder module is depicted in Fig. \ref{fig:encoder}. Specifically, the atom embeddings $A_n^1$ of the first layer are initialized with the atom features $A_n^0$ and the global position embedding $GPE_n$, as described by
\begin{equation}
A_n^1= \mathrm{LayerNorm}\big(\mathrm{Linear}(A_n^0)+GPE_n\big),
\end{equation}
where $\mathrm{LayerNorm}$ and $\mathrm{Linear}$ indicate the layer initialization and linear transformation operations, respectively. The pair embeddings $\mathit{\Phi}_{ij}^{1}$ are initialized with the 3D pair features $\mathit{\Phi}_{ij}^\mathit{3D}$ and the 2D pair features concatenating the connection feature $\mathit{\Phi}^\mathit{SPD}_{ij}$ and the edge feature $\mathit{\Phi}^\mathit{edge}_{ij}$, which is presented as follows:
\begin{equation}
\mathit{\Phi}_{ij}^{1} = \mathrm{Concat}(\mathit{\Phi}_{ij}^\mathit{SPD}, \mathit{\Phi}_{ij}^\mathit{edge})+ \mathit{\Phi}_{ij}^\mathit{3D}.
\end{equation}

Then, the atom and pair embeddings are updated through modified multihead self-attention layers, which build attention weights for each atom and incorporate the current pairwise representation as an additional bias to provide the geometric and spatial information. The equations can be described as follows:
\begin{equation}
\begin{gathered}
Q_n^{l,h}, K_n^{l,h}, V_n^{l,h} = \mathrm{Linear}(A_n^l), \\
M_{ij}^{l,h} = {Q_i^{l,h}(K_j^{l,h})^T}/{\sqrt{d}}  + \mathit{\Phi}_{ij}^{l,h},
\end{gathered}
\end{equation}
where $h \in \{1,...,H\}$ and where $H$ denotes the number of attention heads. $M_{ij}^{l,h}$ represents the attention weight matrix of the $h$-th head on the $l$-th layer, which needs to be refined further to learn the flexible molecule structure information in the ligand encoder. Therefore, a talking-head attention scheme \cite{shazeer2020talking} is leveraged to build a structural understanding of molecules across different modalities, presented as follows:
\begin{equation}
M^{l,h}= \mathrm{softmax} ( M^{l,h} W^{t1}_{l,h})W^{t2}_{l,h},
\end{equation}
where $W_{l,h}^{t1} \in \mathbb{R}^{H \times H}$ and $W_{l,h}^{t2} \in \mathbb{R}^{H \times H}$ are learnable parameters. Finally, the updated atomic representations can be obtained as follows:
\begin{equation}
\label{eq1}
A^{l+1}=\mathrm{LayerNorm} \Big(A^{l} + \mathrm{MLP} \big( \mathrm{Concat}_h(M^{l,h} V_n^{l,h}) W^{l}_O \big)   \Big),
\end{equation}
where $W^{l,h}_O \in \mathbb{R}^{d \times N}$. Simultaneously, the pairwise representations consider the interactive relationships among atoms, which can be updated by concatenating the attention weight matrices directly:
\begin{equation}
\label{eq2}
\mathit{\Phi}_{ij}^{l+1} = \mathrm{Concat}_h\{M_{ij}^{l,h}\}.
\end{equation}

Finally, $L_e$ encoder blocks are stacked to obtain the updated atomic and pairwise representations termed $A^{L_e}$ and $\mathit{\Phi}^{L_e}$. $L_e$ is set to 15 for both encoders in the experiments.

\begin{figure*}[t]
\centering
\includegraphics[width=0.8\linewidth]{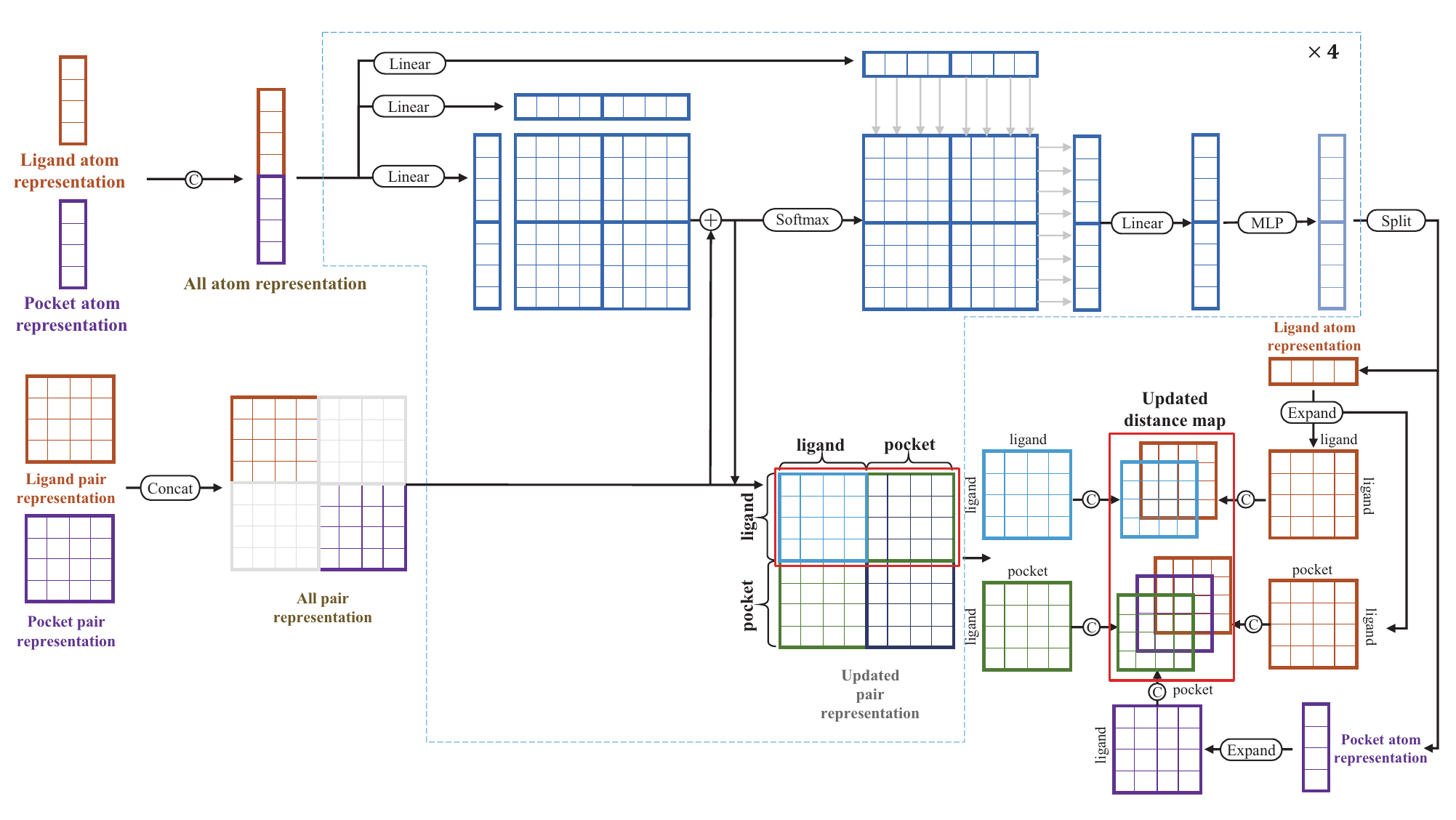}
\caption{Binding module employed to detect interactive relationships between protein pockets and ligands.}
\label{fig:binding}
\end{figure*}

\subsection{Binding Module}
Intramolecular interactions are used to update the atomic and pairwise representations through two separate encoders, whereas intermolecular interactions of atoms between the ligand and protein pocket are taken into account by a binding block. Similarly to the method in \cite{zhou2023unimol}, the binding block employs a similar backbone design with encoders for the sake of simplicity. Fig. \ref{fig:binding} presents the architecture of the binding module. The initialized atomic and pairwise representations of ligand--protein complex, denoted by $C^0$ and $\mathit{\Psi}^0$, are generated by concatenating those of the ligand and protein pocket, described by:
\begin{equation}
\begin{gathered}
C^0 =  \mathrm{Concat}(A_{ligand}^{L_e},  A_{protein}^{L_e}), \\
\mathit{\Psi}^0 =  \mathrm{Concat}(\mathit{\Phi}_{ligand}^{L_e},  \mathit{\Phi}_{protein}^{L_e}), \\
\end{gathered}
\end{equation}
where $C^0$ and $\mathit{\Psi}^0$ are used as inputs of binding blocks, and the padding of $\mathit{\Psi}^0$ is initialized as 0. Similarly, the complex atomic and pairwise representations are updated via Eqs. (\ref{eq1}) and (\ref{eq2}). $C^{L_b}$ and $\mathit{\Psi}^{L_b}$ are achieved through $L_b$ stacked binding blocks, where $L_b$ is set to 4 in the experiments. Finally, the atomic and pairwise representations are disassembled and then reconcatenated to project into the 1-dimensional intra- and intermolecular distance matrices $D^{Intra}_{ij}$ and $D^{Inter}_{ik}$, which are calculated as follows:
\begin{equation}
\label{eq3}
\begin{gathered}
\begin{aligned}
d_{ij}^{Intra} &= W^{Intra}_{1}  \mathrm{LayerNorm}\big(\mathrm{Concat}(C_i^{l},C_j^{p},\mathit{\Psi}_{ij}^{L_b})\big), \\
\bar{d}_{ik}^{Inter} &= \mathrm{RELU} \big( W^{Inter}_1 \mathrm{Concat}(C_k^l, \mathit{\Psi}_{ik}^{L_b}) \big),  \\
D_{ij}^{Intra} &= W^{Intra}_{2} \mathrm{LeakyReLU} (d_{ij}^{Intra}),
\\
\bar{D}_{ik}^{Inter} &= W^{Inter}_2 \mathrm{LayerNorm}(\bar{d}_{ik}^{Inter}), \\
D_{ik}^{Inter} &= (\bar{D}_{ik}^{Inter} + (\bar{D}_{ki}^{Inter})^T)/2,
\end{aligned}
\end{gathered}
\end{equation}
where $i$ and $j$ are the indices of the ligand atoms and $k$ is the index of the atoms in the protein pocket. $W^{Intra}_{1}$, $W^{Intra}_{2}$, $W^{Inter}_{1}$ and $W^{Inter}_{2}$ are learnable parameters.

\begin{figure*}[t]
\centering
\includegraphics[width=0.8\linewidth]{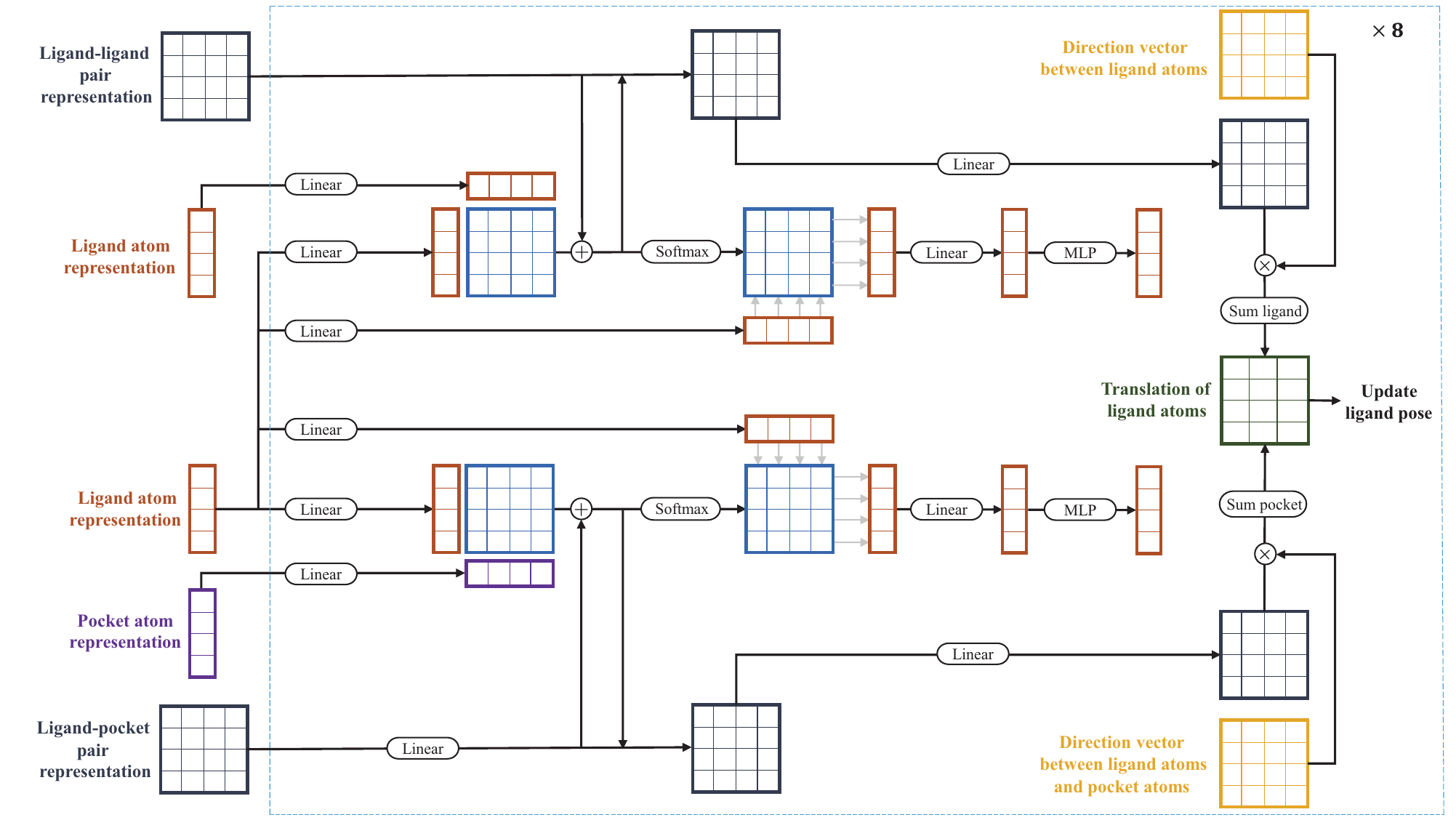}
\caption{Structure module adopted to generate the complex conformations and corresponding confidence assessments.}
\label{fig:st}
\end{figure*}

\subsection{Structure Module}
In previous works, most docking methods, including evolutionary and gradient descent algorithms, use geometry optimization approaches to generate binding conformations. These methods optimize the coordinates of each ligand atom by minimizing the error between the predicted distance matrices and the ground-truth distance matrices. However, these optimization methods are time-consuming and lack robustness because the standalone optimization process needs to be employed for each protein--ligand pair. In addition, the prediction of the binding conformation depends heavily on the precision of the predicted distance matrices, which might introduce more noise. Therefore, Dockformer uses an end-to-end prediction method to generate 3D coordinates of ligand atoms through intra- and intermolecular structure modules, which separately capture potential structural information from ligand-to-ligand and protein-to-ligand interactions, respectively. Specifically, the final complex atomic and pairwise representations are fed into the modules to predict the translation of each ligand atom and update their corresponding coordinates. As illustrated in Fig. \ref{fig:st}, the difference between these modules is that the intramolecular module uses self-attention layers, whereas the intermolecular module adopts cross-attention layers. The atomic and pairwise representations of both ligands and proteins can be updated via Eqs. (\ref{eq1}) and (\ref{eq2}). The coordinates $P_i^l$ of ligand atom $i$ in the $l$-th layer in both modules can be subsequently calculated from the updated representations as follows:
\begin{equation}
\begin{gathered}
a^{intra, l}_{ij} = \mathrm{Linear} \big(\mathrm{Linear}(a^{intra, l-1}_{ij}) + M_{ij}^{intra, l} \big), \\
a^{inter, l}_{ik} = \mathrm{Linear} \big(\mathrm{Linear}(a^{inter, l-1}_{ik}) + M_{ik}^{inter, l} \big), \\
P_i^{l+1} = P_i^{l} +  \sum_{j=1}^n{a^{intra, l}_{ij} \cdot \frac{P_i^l - P_j^l}{||P_i^l - P_j^l||_2}} \\ + \sum_{k=1}^m{a^{inter, l}_{ik} \cdot \frac{P_i^l - P_k^l}{||P_i^l - P_k^l||_2}}, \\
\end{gathered}
\end{equation}
where $a^{intra}$ and $a^{inter}$ are the score matrices and where $M^{intra}$ and $M^{inter}$ denote the attention matrices of the attention layers in both modules. $l \in \{1,...,L_s\}$, where $L_s$ indicates the number of stacked layers and is set to 8 in the experiments.

This end-to-end framework enables Dockformer to directly learn the inherent mapping from input molecular structures to the desired docking conformations. That avoids the need for redundant iterative gradient descent optimization and denoising processes typically required in traditional DL-based docking approaches, significantly reducing computational overhead and greatly saving docking time.

\subsection{Loss Functions}
\label{Lossfunc}
The training process of Dockformer is divided into two phases. First, the outputs of the binding module are mapped to predict the distances between atom pairs. The loss function $\mathcal{L}_{\mathrm{dist}}$ of the predicted intra- and intermolecular distances against the corresponding ground-truth distances can be quantified as follows:
\begin{equation}
\begin{gathered}
\begin{aligned}
\mathcal{L}_{\mathrm{dist}} &= \mathcal{L}_{\mathrm{intradist}} + \mathcal{L}_{\mathrm{interdist}}, \\
\mathcal{L}_{\mathrm{intradist}} &= \frac{1}{2N^2}\sum_{i,j}{(D_{ij}^{Intra}-\hat{D}_{ij}^{Intra})^2}, \\
\mathcal{L}_{\mathrm{interdist}} &= \frac{1}{NM} \sum_{i,k} \mathrm{smooth}_{L1}(D_{ik}^{Inter}-\hat{D}_{ik}^{Inter}), \\
\end{aligned}
\end{gathered}
\end{equation}
where $\hat{D}_{ij}^{Intra}$ and $\hat{D}_{ik}^{Inter}$ denote the ground-truth intramolecular and intermolecular distances, respectively. $N$ and $M$ are the atom numbers of the ligand and protein products, respectively. $i, j \in\{1,...,N\}$ and $k \in \{1,...,M\}$ are the atom indices. $L_2$ loss function is employed for minimizing the intramolecular distance error, whereas a robust $L_1$ loss function $\mathrm{smooth}_{L1}(x)$ is used for minimizing the intermolecular error \cite{girshick2015fast}, presented as:
\begin{equation}
\mathrm{smooth}_{L1}(x) =
\begin{cases}
0.5{x^2}, \quad |x| < 1; \\
x - 0.5, \quad \mathrm{otherwise}.
\end{cases}
\end{equation}

After the first training phase, encoders and binding module can produce effective latent embeddings for both the ligand and protein pockets by considering the interactions between each other. Since the structure module can generate complex conformations in an end-to-end manner, the loss function $\mathcal{L}_{coord}$ with respect to the coordinates of the ligand atoms against the corresponding ground-truth coordinates is incorporated during the second training phase, described by
\begin{equation}
\mathcal{L}_{coord} = \sqrt{\frac{1}{N}\sum_{i=1}^N ( P_i^{L_s} -\hat{P}_i )^2},
\end{equation}
where $P_i^{L_s}$ is the output coordinate of the structure module and where $\hat{P}_i$ denotes the ground-truth coordinate of the $i$-th atom in the cocrystal structure.

In addition, considering that Dockformer is developed to screen small-molecule compounds for a specific target protein virtually, allocating confidence assessment indicators for each generated ligand--protein complex conformation will be constructive. Inspired by the confidence measure in AlphaFold2 \cite{jumper2021highly}, the distance difference test $\mathrm{DDT}_{ij}^{true}$ between the predicted distance $D_{ij}$ and ground-truth distance $\hat{D}_{ij}$ is used to calculate the target confidence of the predicted conformation.
\begin{equation}
\mathrm{DDT}_{ij}^{true} = \frac{100}{4} \sum_{t \in \{0.5, 1, 2, 4\}}
               \frac{\sum_{D_{ij}<8}\mathbf{1}(|D_{ij} - \hat{D}_{ij} < t|)}{\sum_{D_{ij}<8}1},
\end{equation}
where $D_{ij}$ is obtained through the same projection head with atomic representations $C^{L_b}$, $\mathit{t}$ denotes the different thresholds and pairwise representation $\mathit{\Psi}^{L_b}$ in Eq. (\ref{eq3}). The confidence indicator $\mathcal{L}_{\mathrm{conf}}$ can be defined as follows:
\begin{equation}
\begin{gathered}
\begin{aligned}
\mathcal{L}_{\mathrm{conf}} &= \sum_{ij} \mathbf{\hat{p}}^{\mathrm{DDT}}_{ij} \log \mathbf{p}^{\mathrm{DDT}}_{ij} +  \sum_{ik} \mathbf{\hat{p}}^{\mathrm{DDT}}_{ik} \log
\mathbf{p}^{\mathrm{DDT}}_{ik}, \\
\mathbf{\hat{p}}^{\mathrm{DDT}}_{ij} &= \mathrm{onehot}(\mathrm{DDT}_{ij}^{true}), \\
\mathbf{p}^{\mathrm{DDT}}_{ij} &= \mathrm{softmax} \big(\mathrm{MLP}(\mathit{\Psi}_{ij}) \big). \\
\end{aligned}
\end{gathered}
\end{equation}

Finally, the total loss function $\mathcal{L}_{\mathrm{total}}$ of the second training phase can be combined as
\begin{equation}
\mathcal{L}_{\mathrm{total}}=\mathcal{L}_{\mathrm{dist}} + \mathcal{L}_{\mathrm{coord}} + 0.01\mathcal{L}_{\mathrm{conf}}.
\end{equation}

\begin{figure*}[t]
\centering
\includegraphics[width=1\linewidth]{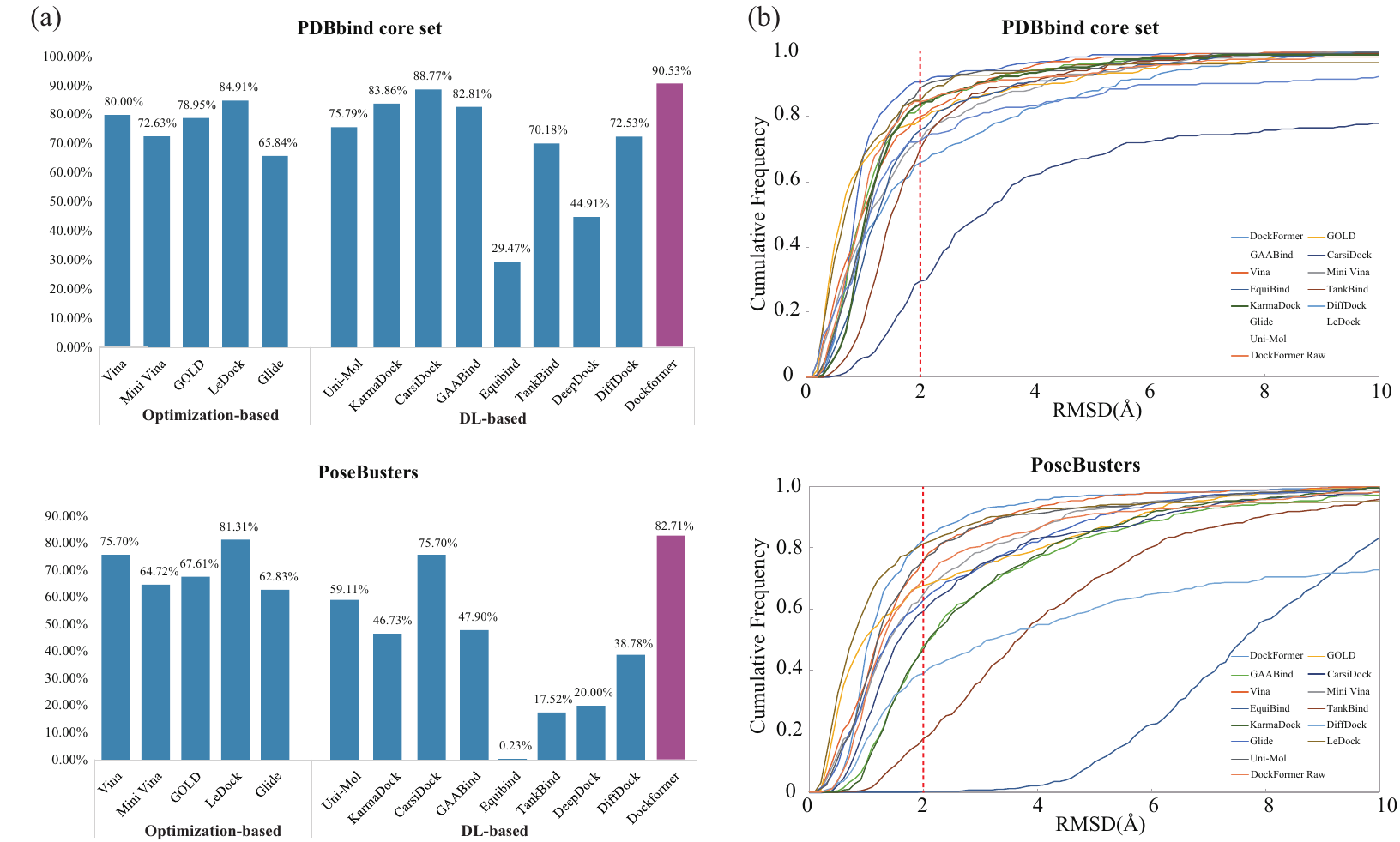}
\caption{Performance comparison of molecular docking methods. (a) Prediction performance of docking methods on the PDBbind core set and PoseBusters dataset. The docking methods are categorized into optimization-based methods and DL-based methods. (b) Cumulative frequency distributions of docking methods on the PDBbind core set and PoseBusters dataset where the $X$-axis is the RMSD threshold, and the $Y$-axis represents the cumulative frequency. The vertical red dashed line is the specified RMSD threshold of $2\mathring{A}$.
}
\label{fig:accuracy}
\end{figure*}

\section{EXPERIMENTS}
\label{experiment}
Extensive experiments are conducted to thoroughly evaluate the effectiveness of Dockformer.
Specifically, Section \ref{Setup} describes the experimental setup, detailing the data preparation, model configurations and evaluation metrics. In Section \ref{Prediction}, the docking accuracy of Dockformer is compared against state-of-the-art methods on Benchmarks. Section \ref{Complexity} explores the computational complexity of Dockformer to highlight how the end-to-end framework accelerates the docking process. In Section \ref{Confidence}, the generated confidence measures are assessed to showcase its ability to reliably distinguish binding strengths, without relying on traditional scoring functions. Section \ref{Screening} demonstrates Dockformer's applicability in an LSVS task to emphasize its scalability and efficiency in real-world drug discovery scenarios. Section \ref{Correction} discusses the implementation of physical plausibility correction, ensuring that the generated conformations are physically realistic and chemically valid. Finally, Section \ref{Ablation} presents the ablation experiments to analyze the contributions of each component to the overall performance and generalization ability of Dockformer, including multimodal fusion and structure module.

\subsection{Experimental Setup}
\label{Setup}
Similarly to most DL-based docking approaches, Dockformer is trained with the latest version of the well-established PDBbind V2020, which includes the cocrystal structures and the corresponding experimentally determined binding affinities of 19443 protein-ligand complexes released before 2020 \cite{liu2015pdb}. The dataset is divided into training and validation sets with a partition ratio of 9:1, using the same filtering protocol provided in Uni-Mol \cite{zhou2023unimol}. In addition, the core set of PDBbind (also termed CASF-2016), which contains 285 hand-curated high-resolution complexes, is used to evaluate the docking ability of Dockformer. Dockformer is further evaluated on an independent test dataset named PoseBusters, which is recently developed and includes 428 crystal complexes released since 2021 \cite{buttenschoen2024posebusters}. By using this dataset, overlap between the training and test datasets is avoided, enhancing the rationality and reliability of the evaluations. In addition, for the proposed Dockformer, the pockets are used as Dockformer's inputs instead of entire proteins to reduce the computational complexity because the binding sites of target proteins have been well studied in most virtual screening tasks. Even without exact pockets, some prediction algorithms can also effectively identify potential binding sites of proteins. Dockformer is trained using eight NVIDIA RTX A6000 GPUs and two 128-core Intel(R) Xeon(R) Gold 6338 CPUs @ 2.00 GHz. In both training phases, the adaptive moment estimation (Adam) optimizer is used to minimize the loss functions smoothly with a learning rate of $10^{-4}$, and an early stopping mechanism based on validation error is employed to prevent overfitting with predefined patience of 20 epochs. In addition, to ensure the fairness of experiments, the configurations of each optimization-based algorithm are maintained according to the default settings, and the parameters of DL-based methods are strictly configured as their original implementation details. That can reduce potential biases introduced by inconsistent parameter tuning and ensure the validity of the performance evaluation results.

\subsection{Prediction Performance on Benchmark Datasets}
\label{Prediction}
In this section, the docking accuracy of Dockformer is evaluated on the PDBbind core set and PoseBusters dataset and compared with that of five commonly used optimization-based docking algorithms, namely, GOLD \cite{verdonk2003improved}, Glide \cite{friesner2004glide}, LeDock \cite{liu2019using}, AutoDock Vina \cite{trott2010autodock} and Mini Vina, and nine state-of-the-art DL-based docking approaches, i.e., DeepDock \cite{mendez2021geometric}, EquiBind \cite{stark2022equibind}, TankBind \cite{lu2022tankbind}, DiffDock \cite{corso2023diffdock}, Uni-Mol \cite{zhou2023unimol}, KarmaDock \cite{zhang2023efficient}, GAABind \cite{tan2024gaabind}, CarsiDock \cite{cai2024carsidock} and Umol \cite{bryant2024structure}.

DL-based docking methods are susceptible to the input training sample distribution, and changing the search space dramatically decreases the model capacity. Therefore, the trained models provided by the official repositories with the default search spaces are used in the comparison study. Specifically, since EquiBind, DiffDock, TankBind and Umol are trained for blind docking tasks, their search spaces cover the entire crystal protein. The binding pockets of Uni-Mol, GAABind and Dockformer are defined as the protein residues within the range of $6\mathring{A}$ from any heavy atom of a crystal ligand, whereas those of KarmaDock and CarsiDock are considered the protein residues within the range of $12\mathring{A}$ and $5\text{-}7\mathring{A}$, respectively. DeepDock considers the protein surface mesh nodes within $10\mathring{A}$ of any crystal ligand atom as inputs. In addition, the number of binding pocket boxes is set to $12\mathring{A}$ for all conventional docking methods. The root mean square deviation (RMSD), which measures the geometric similarity between the predicted binding conformations and the crystal structures of the ligands, is used to evaluate the docking approaches. Generally, binding pose predictions are considered successful when their RMSDs are below the threshold of $2.0 \mathring{A}$ \cite{gaillard2018evaluation}.

As illustrated in Fig. \ref{fig:accuracy}(a), Dockformer achieves the highest docking success rates of 90.53\% and 82.71\% on the PDBbind core set and PoseBusters dataset, respectively. These rates are higher than those of all the baselines. CarsiDock is the second-best model, with slight success rate decreases of 1.76\% and 7.01\% on the two benchmarks, respectively. TankBind, EquiBind and DiffDock perform relatively poor because they are originally trained to solve blind docking tasks. In addition, the docking performance on the PoseBusters dataset is worse for all approaches but especially for DL-based methods. For example, KarmaDock achieves a success rate of 83.86\% using the PDBbind core set and 46.73\% using PoseBusters. This finding implies that some DL-based methods may not generalize well to unseen data because PoseBusters contains only complexes released since 2021. However, the accuracy of Dockformer ranks first on this dataset, suggesting its strong generalizability. Interestingly, most optimization-based docking methods achieve satisfactory and robust performance on both datasets. These methods are superior to those of most DL-based methods. LeDock achieves the second-best accuracy and is slightly inferior to Dockformer using PoseBusters. As a more comprehensive and effective method, the cumulative frequencies of the binding poses with the RMSDs from their corresponding crystal ligands for all the docking methods are plotted in Fig. \ref{fig:accuracy}(b). We find that the success rates of binding poses generated by GOLD and LeDock are higher than those of most DL-based docking approaches. However, Dockformer still performs very competitively with different RMSD cutoffs on both benchmarks, indicating its obvious superiority in terms of accuracy.

\begin{figure}[h!]
\centering
\includegraphics[width=1\linewidth]{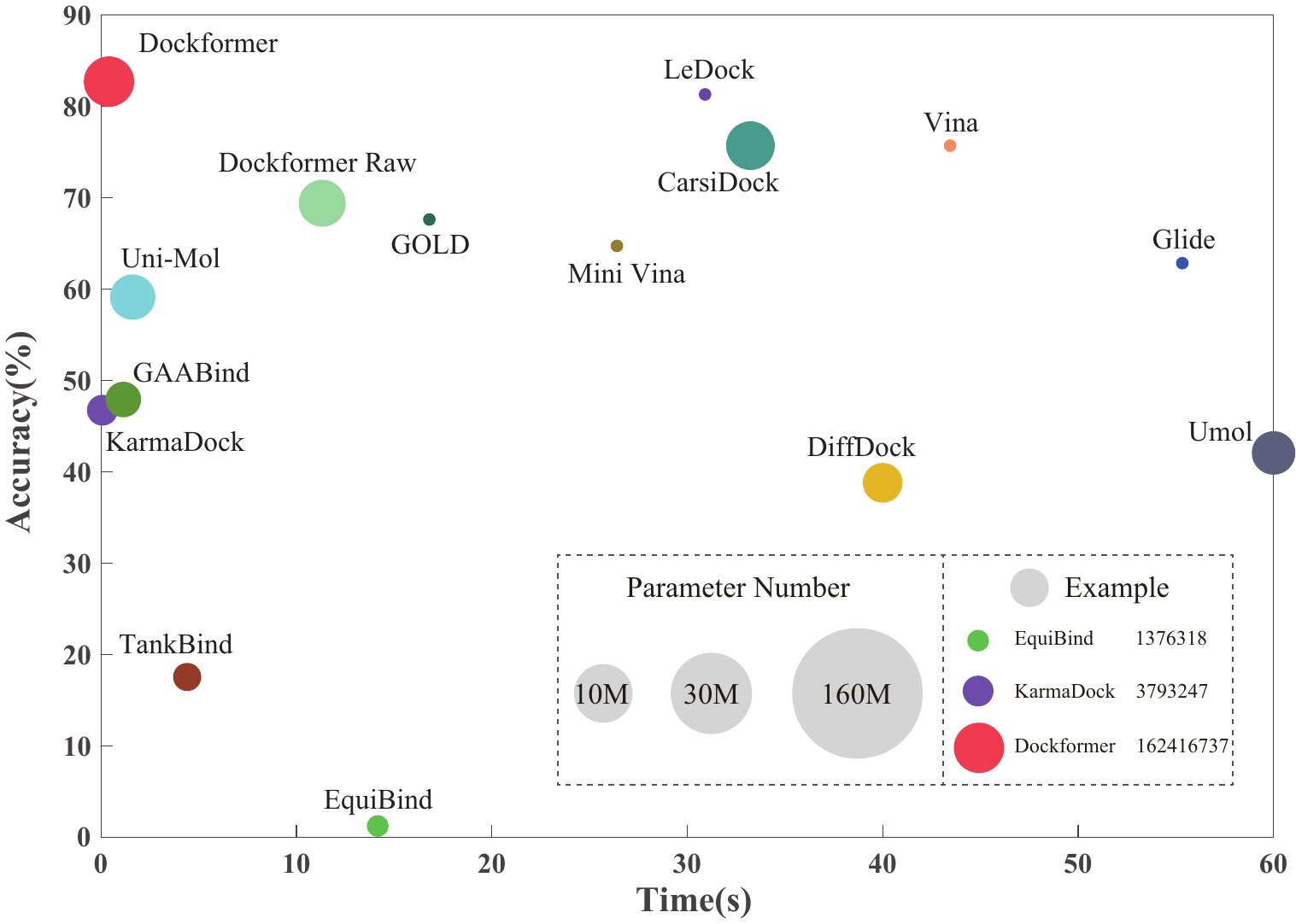}
\caption{Docking accuracy versus inference time of docking approaches using the PoseBusters dataset. The $Y$-axis represents the success rates and the $X$-axis denotes the average inference time of 428 complexes in the benchmark dataset. The size of the circle is proportional to the number of parameters of each docking approach.
}
\label{fig:acc_times}
\end{figure}

\subsection{Computational Complexity versus Accuracy}
\label{Complexity}
In addition to accuracy, computational complexity is an important performance indicator that requires attention, especially when the screening compound library becomes extremely large. Most docking methods cannot traverse the entire library within an acceptable running time. As depicted in Fig. \ref{fig:acc_times}, Dockformer yields the highest docking accuracy and requires the least amount of time using the PoseBusters dataset among all the docking methods. Although Dockformer has the largest DL network, end-to-end binding conformation generation results in low computational complexity in the inference process. Both LeDock and CarsiDock can also achieve competitive docking accuracies but with a long inference time. Thus, these approaches are unsuitable for LSVS tasks. Moreover, most DL-based docking approaches are more efficient than optimization-based approaches, but they sacrifice accuracy. In addition, the docking accuracy of DL-based methods improves as the size of the network architecture increases.

The computational cost of optimization-based docking methods is substantially greater than that of DL-based methods. The adopted stochastic optimization algorithms require more computing resources to determine the position, orientation, and torsion angles of each ligand conformation according to the scoring functions. The DL-based methods, including EquiBind, TankBind, Uni-Mol, GAABind and CarsiDock, use DL algorithms to construct intra- and intermolecular distance maps and then adopt optimization algorithms to calculate the coordinates of ligand atoms for binding pose generation. These methods cannot abandon independent optimization procedures for each ligand, which are faster than those of optimization-based docking methods but still time-consuming because of the iterative gradient descent processes. Only Dockformer and KarmaDock use end-to-end network architecture modules to generate binding conformations in a batch fashion. These modules have prominent advantages in docking efficiency. A modified version of Dockformer termed Dockformer Raw is also evaluated in this experiment. Dockformer Raw uses the same geometry optimization strategy as that used in TankBind to predict binding conformations.

The experimental results demonstrate that Dockformer Raw performs worse in terms of both accuracy and inference time. This finding highlights that the superiority of end-to-end structures can effectively mitigate the computational burden of iterative optimization. Notably, AlphaFold3 achieves a success rate of 90.2\% on the PoseBusters dataset, a rate much higher than that achieved by any of the docking methods mentioned in our experiments. However, even in the case of the fewest number of tokens, the inference time of AlphaFold3 is 22 seconds on 16 A100 graphics processing units (GPUs). This time is much longer than that of most competitors because the model framework of AlphaFold3 is much larger \cite{abramson2024accurate}, and it uses a stepwise denoising method to decode the atom coordinates of the whole complex structure with a diffusion transformer. However, most researchers in academia and industry cannot afford such computational costs.

\subsection{Confidence Assessment}
\label{Confidence}
Most DL-based approaches cannot be applied to virtual screening tasks directly because they can generate only the binding conformations but cannot predict the binding strengths of these conformations. These approaches are usually aided by well-established scoring functions, increasing the computational complexity of screening \cite{cai2024carsidock}. To avoid this disadvantage, DeepDock learns a statistical potential based on the distance likelihood, and KarmaDock trains mixture density networks to learn intermolecular distance distributions as scoring functions. Empirical evidence has demonstrated that such learned scoring functions lead to more powerful screening performance than conventional physics-based methods do \cite{mendez2021geometric, zhang2023efficient}.

\begin{figure}[t]
\centering
\includegraphics[width=0.88\linewidth]{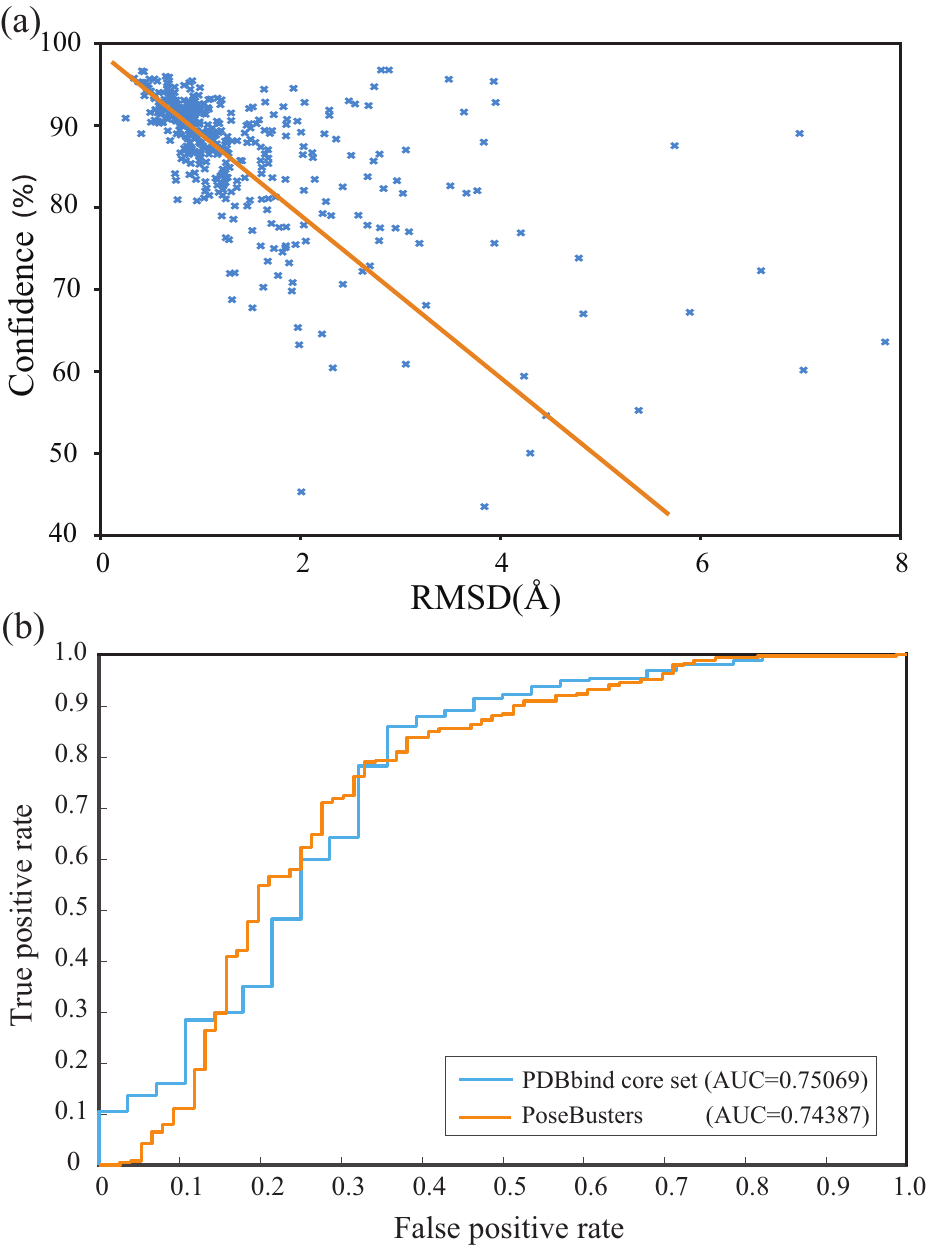}
\caption{Confidence assessment indicators. (a) Distribution of generated binding conformations by Dockformer on benchmark datasets. The $X$-axis denotes the RMSD threshold, and the $Y$-axis represents the confidence of each predicted conformation. (b) ROC curves of confidence assessment indicators for distinguishing whether the predicted conformations are successful on benchmark datasets.
}
\label{fig:2}
\end{figure}

Similarly, Dockformer allocates confidence assessment indicators for each generated complex conformation following the protocol of AlphaFold2. The target confidence of the predicted conformation is estimated via the distance difference test between the predicted and ground-truth distances and can be used to describe the binding strengths between proteins and ligands for virtual screening, which is described in Section \ref{Lossfunc}. To verify the effectiveness of the confidence measures, a scatter diagram of the generated complex conformations with the corresponding confidence indicators and RMSD values is illustrated in Fig. \ref{fig:2}(a). We find an apparent linear relationship between confidence and RMSD, represented by the orange line, through a simple linear regression method. The results suggest that higher confidence indicates lower RMSDs of the generated conformations.

In addition, the effectiveness of the confidence indicators is verified by distinguishing the strong and weak binders. The predicted binding poses are allocated positive or negative labels, depending on whether their RMSDs are above or below the threshold of $2.0 \mathring{A}$. The confidence indicators are subsequently used to classify these conformations, and the receiver operating characteristic curves are presented in Fig. \ref{fig:2}(b). Areas under the curve (AUCs) of 0.7506 and 0.7438 are achieved on two benchmark datasets. These values are much larger than 0.5000, indicating the powerful classification performance of the confidence indicators. Therefore, on the basis of these confidence indicators, Dockformer can be applied to large-scale virtual screening tasks without additional scoring functions.

\subsection{Large-scale Virtual Screening Task}
\label{Screening}
\begin{figure}[ht]
\centering
\includegraphics[width=1\linewidth]{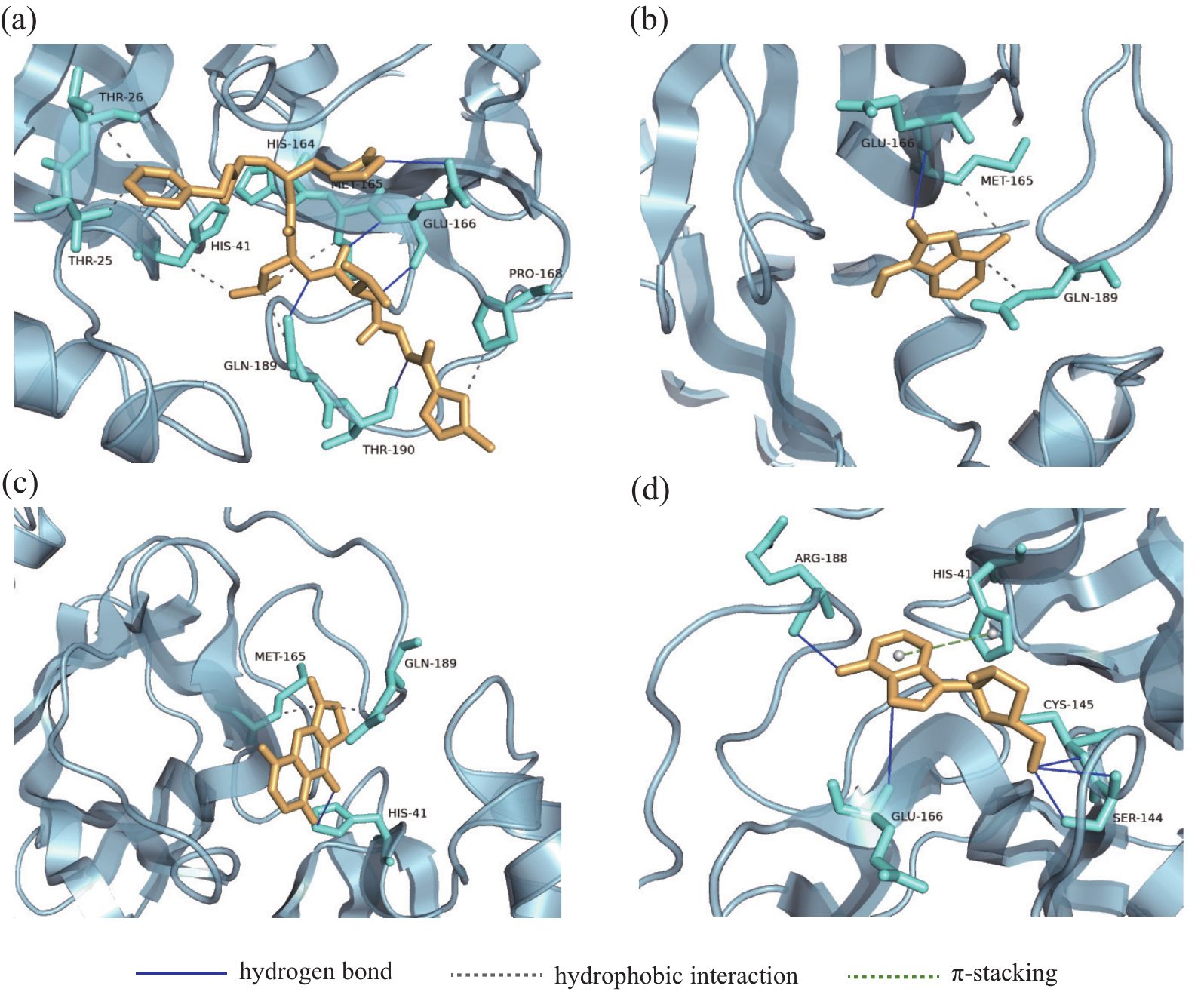}
\caption{Visualization of nonbonded interactions between $\mathrm{M_{pro}}$ and various compounds. (a)-(d) Nonbonding interaction mapping; these panels illustrate the nonbonded interactions between the main protease ($\mathrm{M_{pro}}$) and four different compounds: N3 (a), CHEMBL1571559 (b), CHEMBL277134 (c) and CHEMBL277789 (d). The target protein is depicted in light blue, and the compounds are shown in orange.
}
\label{fig:vs_pose}
\end{figure}

In this section, Dockformer is applied to a real-world virtual screening scenario to verify its screening power. Considering that COVID-19 has challenged economic and healthcare systems worldwide, Dockformer is utilized to screen potential drug candidates for this disease with high transmission and mortality rates. The main protease $\mathrm{M_{pro}}$, whose binding site is highly conserved among all coronaviruses, is selected as the target protein and can serve as a drug target for the design of broad-spectrum inhibitors \cite{inhibitors}. Previous studies revealed that the Michael acceptor inhibitor $\mathrm{N3}$ can specifically inhibit the $\mathrm{M_{pro}}$ of multiple coronaviruses, including SARS-CoV and MERS-CoV, and has shown potent antiviral activity against infectious bronchitis virus in animal models \cite{mpro, mpro2}. The binding pose of $\mathrm{M_{pro}}$ and N3 is illustrated in Fig. \ref{fig:vs_pose}(a). Hydrophobic interactions evidently exist between the residues THR25, MET165, HIS41, and GLN189 and the inhibitor N3, and hydrogen bonds are formed between the residues THR25, GLU166, HIS41, and N3. A large-scale bioactivity database named ChEMBL is used for screening \cite{chembl}. This database contains more than 1.2 million compounds after filtering the molecules whose molecular weights are greater than 400 $g/mol$. Traditional optimization-based docking approaches require more than one year to screen all the compounds, whereas Dockformer completes such screening tasks in less than 48 hours, highlighting its high efficiency.

Table \ref{vs} presents the top 20 compounds with the highest confidence. The compound named CHEMBL1571559 ranks first, and it forms hydrogen bonds with the residue GLU166 and hydrophobic interactions with the residues MET165 and GLN189 in $\mathrm{M_{pro}}$, as shown in Fig. \ref{fig:vs_pose}(b). CHEMBL1571559 and N3 exhibit the same interaction patterns. Similar observations can also be found for the compound named CHEMBL277134 in Fig. \ref{fig:vs_pose}(c). This compound forms hydrogen bonds with residue HIS41 and hydrophobic interactions with residues MET165 and GLN189. In addition, a well-characterized serotonin antagonist named cinanserin can inhibit SARS $\mathrm{M_{pro}}$ by forming cation--$\pi$ interactions with the benzene rings of the residues HIS41 and GLU166 \cite{cinanserin}, \cite{cinanserin2}. The same interactions can also be observed for the compound named CHEMBL277789, which not only forms hydrogen bonds with the residues ARG188, GLU166, CYS145, and SER144 but also has $\pi$ stacking with the ring structure of the residue HIS41, as shown in Fig. \ref{fig:vs_pose}(d).

To sum up, these results validate the potential screening power of Dockformer in practical applications, considering it can efficiently screen a large-scale molecular library and precisely identify candidate compounds similar to the known inhibitor N3. That showcases its ability to find potential drug candidates within a vast chemical space quickly and accurately, underscoring its potential utility in accelerating drug discovery efforts.

\begin{table}[t]
\caption{Top-20 compounds virtually screened by Dockformer binding to $\mathrm{M_{pro}}$}
\centering
\scriptsize
\begin{tabular}{lccc}
\hline \hline
Name & PubChem ID & Confidence (\%) & Weight ($g/mol$)  \\ \hline

CHEMBL1571559 & 600593 & 97.0395 & 176.17  \\
CHEMBL1495267 & 839273 & 96.9385 & 176.17  \\
CHEMBL1866947 & 750687 & 96.9329 & 180.14  \\
CHEMBL1331002 & 741419 & 96.8854 & 175.25  \\
CHEMBL1471518 & 601953 & 96.8728 & 180.14  \\
CHEMBL22608 & 10726577 & 96.8647 & 196.59  \\
CHEMBL373066 & 44407700 & 96.8593 & 186.17  \\
CHEMBL8130 & 136023340 & 96.8491 & 201.18  \\
CHEMBL1709155 & 135447995 & 96.8387 & 199.21  \\
CHEMBL1980161 & 20135774 & 96.8339 & 167.59  \\
CHEMBL1372588 & 11401613 & 96.8286 & 213.66  \\
CHEMBL312138 & 516636 & 96.8281 & 146.15  \\
CHEMBL8362 & 135429981 & 96.8223 & 215.21  \\
CHEMBL1444165 & 5418133 & 96.8163 & 221.62  \\
CHEMBL1885120 & 242567 & 96.8147 & 144.17  \\
CHEMBL5281891 & 11117054 & 96.8139 & 200.19  \\
CHEMBL4542958 & 28342906 & 96.8049 & 149.15  \\
CHEMBL1993673 & 135489792 & 96.7984 & 204.21  \\
CHEMBL1836358 & 10104270 & 96.7928 & 201.25  \\
CHEMBL277134 & 44269793 & 96.7886 & 249.25  \\
\hline \hline
\label{vs}
\end{tabular}
\end{table}

\subsection{Physical Plausibility Correction}
\label{Correction}
The major disadvantage of DL-based docking methods, including Dockformer, is that their predicted complex conformations might lack physical plausibility. These algorithms directly generate the 3D coordinates of each ligand atom instead of the translation, orientation and torsion of the ligands, intentionally increasing the degree of freedom of docking problems but inevitably distorting the inherent topological structures of the molecules. To address this issue, three postprocessing methods—point cloud fitting-based alignment (PCF), force-field optimization (FF), and energy minimization (EM)—are proposed to refine the predicted binding conformations. Specifically, PCF uses a distance geometry-based method to apply the transformations to the generated conformation. FF employs an iterative process to optimize the conformations regardless of the protein pocket structure. Thus, it fails to guarantee intermolecular validity, generally leading to overlap and steric hindrance between ligand and protein atoms. EM modifies the ligand poses by minimizing the binding energy of the ligand--protein complex, considering the rigid structure of the protein pocket.

\begin{figure*}[ht]
\centering
\includegraphics[width=1\linewidth]{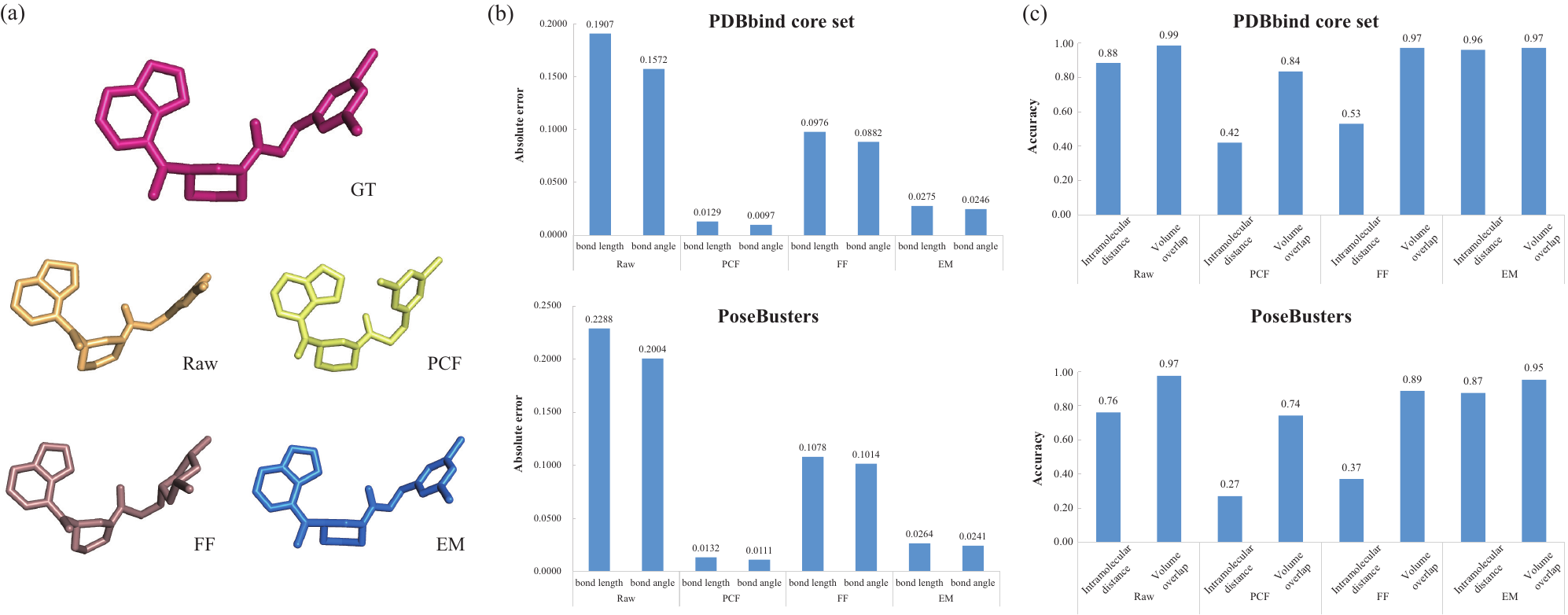}
\caption{Comparison of different postprocessing methods. (a) Visualization of binding conformations revised by different postprocessing methods. (b) Absolute errors in the bond length and bond angle for different postprocessing algorithms on the PDBbind core set and PoseBusters dataset. These charts highlight the effectiveness of different postprocessing algorithms in minimizing the absolute errors in the bond length and bond angle, providing insights into their performance on the PDBbind core set and PoseBusters dataset. (c) Accuracy of different postprocessing algorithms on intramolecular distance and volume overlap for the PDBbind core set and PoseBusters dataset. These charts illustrate the effectiveness of different postprocessing algorithms in accurately predicting intramolecular distance and volume overlap, providing insights into their performance on the PDBbind core set and PoseBusters dataset.}
\label{fig:3}
\end{figure*}

Fig. \ref{fig:3}(a) shows a visualization of binding poses modified by different methods. There are apparent topological distortions in the raw predicted conformation compared with the ground-truth structure. FF fails to correct incorrect topological structures. Although PCF provides the correct local structures, it dramatically changes the original torsion of the ligand, influencing the intermolecular interactions between the ligand and the protein pocket. EM can provide a physically plausible conformation and maintain the original translation, orientation and torsions of the ligands as much as possible. Figs. \ref{fig:3}(b) and \ref{fig:3}(c) present the performance of these algorithms on both the PDBbind core set and PoseBusters dataset.

The binding poses predicted by Dockformer lead to bond length errors of 0.1907 and 0.2288 and angle errors of 0.1572 and 0.2004 on the PDBbind core set and PoseBusters dataset, respectively. PCF obtaines the lowest errors for both measures among the postprocessing methods. FF produces unreasonable local structures, leading to higher errors than the other strategies did. Regarding protein--ligand distance restrictions and volume overlaps, the predicted binding modes from Dockformer achieve correct rates of 0.88 and 0.99 on the PDBbind core set, and 0.76 and 0.97 on the PoseBusters dataset. However, these two measures decrease significantly after the PCF and FF strategies are applied, especially for distance restrictions. The EM method maintains the highest accuracy on both benchmarks since the rigid structure of the protein pocket is a mandatory requirement to optimize the ligand conformations. In summary, the EM strategy can effectively refine the binding poses predicted by Dockformer to guarantee physical plausibility.

\begin{table}[t]
\caption{Ablation Study of Docking Success Rate and RMSD on the PDBbind core set}
\label{tab1}
\centering
\scriptsize
\begin{tabular}{lcccc}
\hline \hline
Model & Success rate (\%) & $\textrm{RMSD} (\mathring{A})$ \\ \hline

Dockformer & \textbf{90.53} & \textbf{1.14} \\
Dockformer Raw & 84.91 & 1.43 \\
Without global position embeddings & 76.49 & 1.78 \\
Without multimodality information & 84.56 & 1.38 \\
Without talking-head attention & 85.96 & 1.26 \\
Without update representation by decoder & 85.61 & 1.25 \\
Larger binding site & 74.04 & 1.82 \\
\hline \hline
\end{tabular}
\end{table}

\subsection{Ablation Studies}
\label{Ablation}
In this section, ablation experiments are conducted to estimate the importance of each component of Dockformer, including the structure module, multimodality information, binding site size, and number of rotatable bonds. Finally, the conclusions of this study and future works are presented.

\begin{figure*}[t]
\centering
\includegraphics[width=0.8\linewidth]{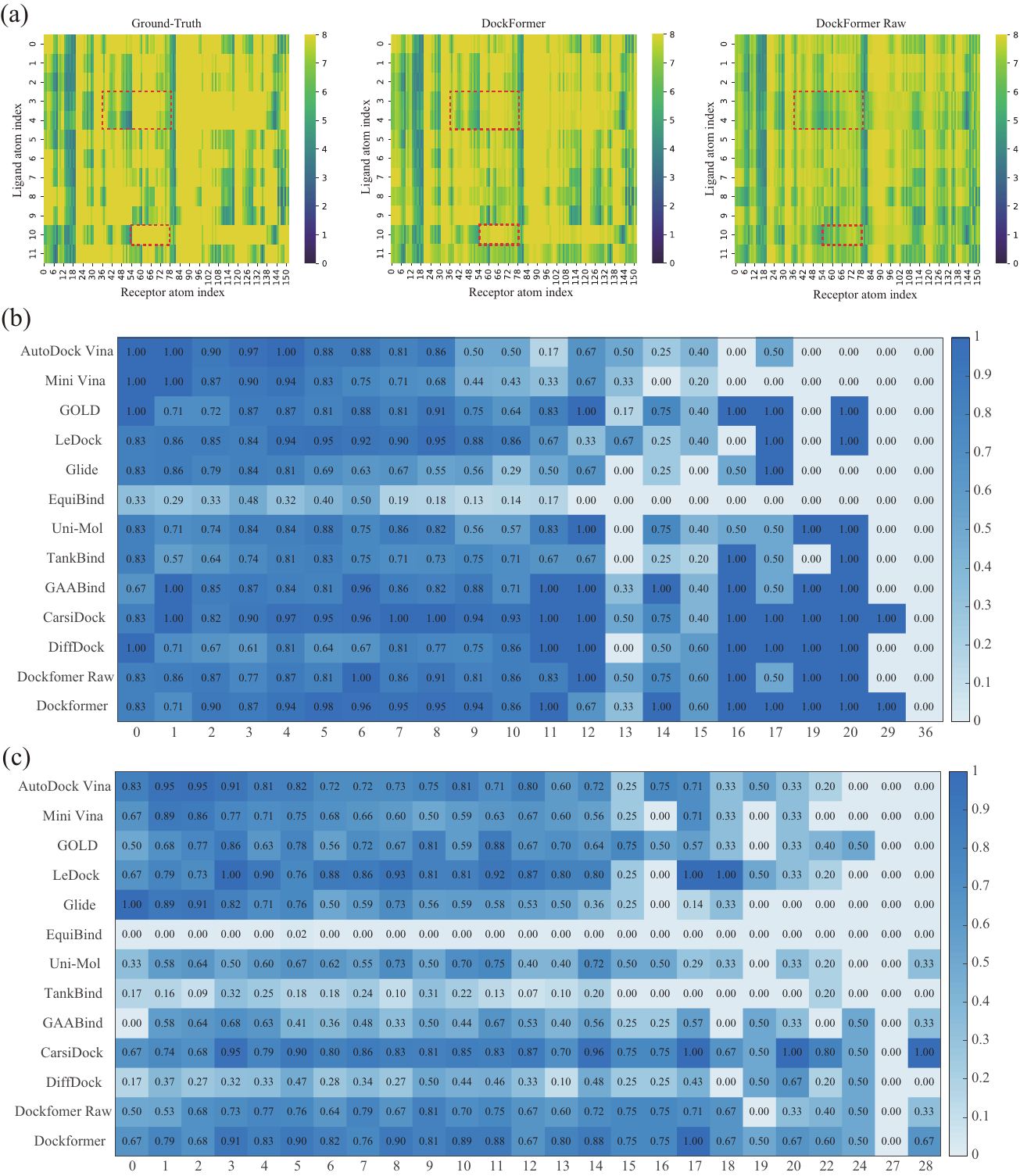}
\caption{Improvement of the decoder and impact of the number of rotatable bonds. (a) Predicted distance distributions between ligand and protein atoms for the ground-truth, Dockformer, and Dockformer Raw. (b),(c) Performance of traditional docking methods and Dockformer among different dimensionalities of the docking problem on the PDBbind core set (b) and PoseBusters dataset (c).}
\label{fig:5}
\end{figure*}

\textit{1) Impact of the Structure Module:} To prove the effectiveness of the structure module, the performance of Dockformer Raw is evaluated and compared with that of the baseline model. Dockformer Raw uses a gradient descent method to generate the binding pose instead of the structure module, which is based on the predicted distance matrices. As presented in Table \ref{tab1}, Dockformer Raw achieves a success rate of 84.91\% and an RMSD of $1.43\mathring{A}$ on the PDBbind core set. These values are much lower than those of Dockformer. In addition, the gradient descent method updates the ligand coordinates iteratively, leading to more time consumption during optimization. As shown in Fig. \ref{fig:acc_times}, Dockformer Raw requires 11.32 seconds per ligand, which implies much greater computational complexity than Dockformer does. The predicted distance distributions between ligand and protein atoms derived from the pairwise representations of Dockformer and Dockformer Raw are compared in Fig. \ref{fig:5}(a). Compared with the ground-truth distance map, Dockformer Raw fails to characterize specific interactions, such as those between ligand atoms 3-4 and receptor atoms 36-78, as well as the interaction between ligand atom 10 and receptor atoms 54-78. These results suggest that the structure module can capture more potential interactive knowledge to generate accurate binding poses than regular optimization algorithms can.

\textit{2) Impact of Multimodality Information: }Compared with the previous transformer-based docking approaches, Dockformer uses multimodal molecular information to enhance the model capabilities via the following strategies. First, as the conventional self-attention mechanism lacks positional information, the positional embeddings for the 1D sequence information are added to each token in the natural language process field. Similarly, learnable global position embeddings, which reflect the spatial information of 3D atomic coordinates, are added to atom embeddings for both ligands and proteins in Dockformer. As expected, the success rate and RMSD of the main architecture without global position embeddings remarkably decrease compared with those of the baseline model (76.49\% $vs.$ 90.53\% and 1.78$\mathring{A}$ $vs.$ 1.14$\mathring{A}$, respectively), implying the effectiveness of global position embeddings. Second, molecular structures contain inherent topological structures and local rigid fragments, which can be interpreted as both 2D graphs and 3D point clouds. To fully use the multimodal information, Dockformer concatenates the 2D graph information of connection and bond features and the 3D geometric information of interatomic distance features to generate pair representations. As shown in Table \ref{tab1}, without using multimodal information, the docking success rate and RMSD of the Dockformer decrease to 84.56\% and $1.38\mathring{A}$, respectively. In addition, the multimodal representations are then incorporated to update the atom embeddings as the bias of the talking-head attention mechanism. Removing the talking-head attention changes the success rate and RMSD to 85.96\% and $1.26\mathring{A}$ respectively, emphasizing the great importance of multimodality information in Dockformer.

\textit{3) Impact of the Pocket Size: }
For optimization-based docking algorithms, the binding site size directly determines the computational complexity of molecular docking tasks. Specifically, a larger binding site leads to a more complicated search process, whereas a smaller site may result in a ligand beyond the search scope and an unsuccessful docking process. Some DL-based algorithms, including EquiBind and TankBind, confirm binding by considering the whole protein structure. These proteins are usually characterized on the basis of residues rather than atoms to reduce the computational complexity of models. However, such coarse-grained representations limit the models' generalizability, making them suitable for blind docking experiments. Very recently, more pocket-centralized models, such as Uni-mol, GAABind and CarsiDock, have been reported. The binding box of the original Dockformer is set to $6\mathring{A}$, similar to other DL-based approaches. In this section, the performance of the Dockformer is evaluated with a larger box size of $10\mathring{A}$. Dockformer achieves a success rate of 74.04\% and an average RMSD of $1.82\mathring{A}$, which are much worse than those of the baseline model. These results are obtained because a larger pocket requires these DL-based methods to explore inherent intermolecular interactions from a greater number of protein--ligand atom pairs, which usually rely on larger network architectures and more available training samples.

\textit{4) Impact of the Number of Rotatable Bonds: }
The number of rotatable bonds in a ligand directly determines the difficulty of solving molecular docking problems. There is a general consensus that optimization-based methods suffer from the curse of dimensionality, and the search abilities of optimization algorithms deteriorate with increasing number of rotatable bonds \cite{ji2022autodock}. The heat maps of the success rates of the docking methods for different numbers of rotatable bonds are presented in Fig. \ref{fig:5}(b). It is apparent that the success rates of most traditional docking programs significantly decrease when the number of rotatable bonds increases, and satisfactory predictions are obtained only for compounds with fewer than 10 rotatable ligand bonds. However, Dockformer can maintain reliable docking accuracy, disregarding the flexibility of molecules, which verifies its robust and generalized performance. Fig. \ref{fig:5}(b) shows the general advantage of most DL-based approaches because larger compounds enable DL models to learn more deterministic atom pair interactions between ligands and pockets, which may contribute to the development of molecular drugs with heavier weights.

\section{CONCLUSION}
\label{conclusion}
This study proposes Dockformer for LSVS, which integrates multimodal fusion, positional encoding and end-to-end architecture. Compared with conventional DL-based algorithms, these advanced components enable Dockformer to improve docking accuracy and accelerate the docking process significantly. In addition, Dockformer showcases its potential to expedite drug discovery efforts by efficiently screening large molecular libraries and precisely identifying compounds with similar interaction patterns to known inhibitors. As a robust and reliable protein-ligand docking approach, Dockformer holds promise for significantly reducing the development cycle and cost of drug design.

With the aid of artificial intelligence technologies, large-scale compound libraries can be explored to pursue promising drug candidates with higher diversity and stronger binding strengths, which may improve the hit rates to target specific proteins of interest. Following this perspective, an \textit{in silico} molecular docking algorithm named Dockformer, whose end-to-end framework enables improvements in docking accuracy and screen efficiency simultaneously, is proposed in this study. However, although existing computational methods have undergone tremendous advances, detecting the biological interactions between proteins and ligands remains challenging because of the scarcity of available training data, physical implausibility, and false-positives of hit identifications. In addition, the Chemical Universe Databases will contain trillions of compounds in the coming years, requiring more efficient high-performance screening methods to search such extremely large regions of the chemical space. \textit{De novo} drug design methods may be an alternative to docking algorithms to skip the computationally expensive screening process on the basis of generative models and deep reinforcement learning approaches. The source code of Dockformer is available at \url{https://zenodo.org/records/12792385}.

\bibliographystyle{IEEEtran}
\bibliography{sn-bibliography}

\end{document}